\documentclass[12pt]{article}

\usepackage[utf8]{inputenc}
\usepackage[T1]{fontenc}
\usepackage[english]{babel}

\usepackage[margin=1in]{geometry}
\usepackage[expansion=false]{microtype}

\usepackage{amsmath,amssymb}
\usepackage{lmodern}

\usepackage{caption}
\usepackage{setspace}
\newlength{\parbreakskip}
\newcommand{\parbreak}{\\[\parbreakskip]}
\usepackage{graphicx}
\usepackage{booktabs}
\usepackage{array}
\usepackage{enumitem}
\usepackage{xcolor}
\usepackage{tikz}
\usetikzlibrary{positioning, arrows.meta, shapes.geometric}

\usepackage{listings}
\definecolor{vscbackground}{rgb}{0.9686,0.9686,0.9686}
\definecolor{vsccomment}{HTML}{008000}
\definecolor{vsckeyword}{HTML}{0000FF}
\definecolor{vscstring}{HTML}{A31515}
\definecolor{vsclinenumber}{rgb}{0.549,0.549,0.549}
\lstdefinestyle{mystyle}{
  language=Python,
  morekeywords={None,True,False,self},
  commentstyle=\color{vsccomment},
  keywordstyle=\color{vsckeyword}\bfseries,
  stringstyle=\color{vscstring},
  numberstyle=\tiny\color{vsclinenumber},
  basicstyle=\ttfamily\small,
  breakatwhitespace=false,
  breaklines=true,
  captionpos=b,
  keepspaces=true,
  numbers=left,
  numbersep=8pt,
  showspaces=false,
  showstringspaces=false,
  showtabs=false,
  tabsize=4,
  columns=fullflexible,
}
\usepackage{natbib}
\usepackage{bibentry}
\usepackage[colorlinks=true, linkcolor=blue!50!black, citecolor=blue!50!black, urlcolor=blue!50!black]{hyperref}

\usepackage{titlesec}
\titleformat{\section}{\normalfont\Large\bfseries\sffamily}{\thesection.}{0.25em}{}
\titleformat{\subsection}{\normalfont\large\bfseries\sffamily}{\thesubsection.}{0.25em}{}
\titleformat{\subsubsection}{\normalfont\normalsize\bfseries\sffamily}{\thesubsubsection.}{0.25em}{}
\titlespacing*{\section}{0pt}{1.4em}{0.6em}
\titlespacing*{\subsection}{0pt}{1.1em}{0.4em}

\usepackage{fancyhdr}
\usepackage{float}
\usepackage{placeins}
\usepackage{caption}
\usepackage{subcaption}

\newcommand{\authorentry}[4]{%
  \begin{minipage}[t]{\authorboxwidth}
    \centering
    \mbox{\textbf{#1}\textsuperscript{#2}}\\[3pt]
    \texttt{#3}\\
    ORCID: \href{https://orcid.org/#4}{#4}
  \end{minipage}%
}
\newlength{\authorboxwidth}
\begin{document}
\nobibliography*

\begin{center}
  {\sffamily\LARGE\bfseries Technical Manual for a Toolkit for Measuring Contextual Individuation in Transformer Language Models}
\end{center}

\vspace{10pt}
\hrule
\vspace{10pt}

\begin{center}
\footnotesize
\authorentry{José Luciano Verçosa Marques}{1}{zlvm@unicamp.br}{0000-0002-3191-0846}%
\hspace{0.015\textwidth}%
\authorentry{Frederico Jorge Heitmann}{2}{f023863@dac.unicamp.br}{0009-0006-5828-8301}%
\hspace{0.015\textwidth}%
\authorentry{Daniel Omar Perez}{3}{doperez@unicamp.br}{0000-0002-5965-3490}

\vspace{16pt}

\authorentry{Marcelo Vinicius de Paula}{1}{mvpaula@unicamp.br}{0000-0002-2213-6086}%
\hspace{0.015\textwidth}%
\authorentry{Tárcio André dos Santos Barros}{1}{tarcio87@unicamp.br}{0000-0001-9413-1279}
\end{center}

\vspace{8pt}
\begin{flushleft}
\scriptsize
\textsuperscript{1}~Center for Electric Mobility Research (CEMOBE) / Power Electronics Laboratories (LEPO), University of Campinas (Unicamp)\\
\textsuperscript{2}~Institute of Computing (IC), University of Campinas (Unicamp)\\
\textsuperscript{3}~Center for Logic, Epistemology and History of Science (CLE), University of Campinas (Unicamp)
\end{flushleft}

\vspace{1pt}
\begin{center}
\scriptsize Corresponding author: José Luciano Verçosa Marques (\texttt{zlvm@unicamp.br})
\end{center}
\normalsize

\vspace{6pt}
\hrule
\vspace{6pt}

\begin{abstract}
\noindent
A transformer language model assigns a single, context-independent vector to a word type at its embedding layer, yet is widely believed to individuate that word's occurrences by context in its later layers. Testing this belief cleanly requires a construct that holds the word \emph{form} fixed while its context and intended sense vary in a controlled, labeled way. This manual documents an open toolkit built around such a construct, which we call a \emph{bridge form}: a single written word that recurs, unchanged, across two or more subject domains with a different sense in each. We describe, and justify, every stage of the pipeline: the declarative specification of bridge forms and their source domains, corpus acquisition from Wikipedia, occurrence localization, layer-wise representation extraction, a domain-pairwise silhouette measurement of separation in the model's representation space, and a paired visualization protocol. Each design choice is presented together with the methodological failure mode it is meant to avoid (sense contamination from overly broad category labels, the multi-group bias of the silhouette coefficient, subword-tokenization misalignment, and axis-comparability artifacts in dimensionality-reduced plots, among others). This manuscript is a methodological and implementation reference: it does not report or interpret empirical outcomes of running the toolkit on any particular model or bridge-form set. The toolkit, its full source, and the corpora used to exercise it are archived separately (Section~\ref{sec:availability}) under a persistent identifier, and are intended to be cited as an instrument by studies that use it to produce and interpret empirical results.
\end{abstract}

\setstretch{1.5}
\section{Introduction}
\label{sec:intro}
Contextual language models replaced static word embeddings on the premise that a word's representation should depend on where it occurs, not only on which word it is. That premise is easy to state and hard to test cleanly. Most existing evidence for it is indirect: downstream task performance improves when contextual representations are used in place of static ones \citep{peters2018elmo}, or probing classifiers show that certain properties of a sentence are linearly recoverable from a token's hidden state at some layer \citep{tenney2019bert, belinkov2019analysis}. Such evidence is compatible with context-sensitivity but does not isolate it: a probe trained on annotated data can succeed for reasons that have little to do with a model's spontaneous tendency to keep distinct senses of the same word type apart in its own geometry.\parbreak
This manual documents a toolkit built to isolate that specific question directly. The construct it relies on, which we call a \emph{bridge form}, is a single written word that occurs, unchanged, across two or more subject domains, carrying a domain-appropriate sense in each: for instance, \emph{current} as \emph{electric current} (physics), a \emph{current account} (economics), and an \emph{ocean current} (physical geography). Because every occurrence of a bridge form shares the same entry in a model's vocabulary, the model's static embedding-layer output is, up to positional variation, the same vector for all of them. Any separation between domains that appears in later layers can therefore only be a fact about context, not about word identity. This is a structural guarantee that does not depend on trusting the model's downstream behavior or on any annotated sense inventory.\parbreak
The toolkit operationalizes this idea end to end: it declares bridge forms and their domains against specific, non-overlapping source categories; builds a corpus from those categories; locates every occurrence of each form; extracts its hidden-state vector at every layer of a chosen transformer; measures domain separation with a pairwise silhouette coefficient computed at full representation width; and visualizes the same occurrences with a paired, axis-comparable projection. Every one of these stages embeds a methodological choice that is not forced by the problem statement, and each choice can silently determine the result if left implicit. The purpose of this manual is to make every one of those choices explicit, and to justify it, so that the toolkit can be reused, audited, and extended without having to reverse-engineer its design from source code alone.\parbreak
This manuscript is deliberately scoped as a methodological and implementation reference. It does not present, tabulate, or interpret the empirical outcome of running the toolkit against any specific model or bridge-form inventory; it documents an instrument, not a finding.\parbreak
The remainder of this manual is organized as follows. Section~\ref{sec:uses} states what kind of questions the toolkit is intended to help answer. Section~\ref{sec:background} situates the bridge-form construct relative to static and contextual embedding literatures and to existing probing methodology. Section~\ref{sec:design} states and justifies the toolkit's design principles. Section~\ref{sec:toolkit} describes the pipeline stage by stage. Section~\ref{sec:implementation} covers implementation and reproducibility. Section~\ref{sec:interpreting} explains how the toolkit's output is meant to be read, illustrated with example output. Section~\ref{sec:limitations} states the toolkit's scope and known limitations. Section~\ref{sec:availability} gives availability and citation information.

\section{Intended Uses and Potential Contributions}
\label{sec:uses}
The toolkit is an instrument, not a claim, and its value lies in the kinds of questions it makes tractable rather than in any specific answer it has been used to produce. Four uses motivated its design.
\begin{itemize}
    \item \textbf{A training-free complement to probing.} Probing classifiers ask what \emph{can} be linearly decoded from a representation once a decoder is trained on labeled data; that is a different question from whether a model's own geometry already keeps senses apart without any decoder in the loop. The pairwise silhouette measurement of Section~\ref{sec:pipeline-measure} answers the second question directly, on the model's own representation space, with no classifier to train, tune, or trust.
    \item \textbf{Architecture-neutral model comparison.} Because the pipeline treats a model as a black box that returns per-layer hidden states (Section~\ref{sec:design-architecture}), the same instrument applies unmodified to encoders and decoders, small and large, across pretraining objectives. This gives a common, repeatable basis for asking how differently trained or differently shaped models organize context-dependent meaning, without privileging any one architecture as the reference case.
    \item \textbf{An empirical handle for lexical semantics.} Questions about polysemy, homonymy, and the individuation of word senses have traditionally been argued from intuition, from curated dictionaries, or from manually sense-tagged corpora. The bridge-form construct offers a further, complementary source of evidence: whether, and to what extent, a specific computational system that was never given a sense inventory nevertheless organizes a word's occurrences by something resembling sense, purely as a function of context. This does not settle any philosophical question on its own, but it supplies a concrete, reproducible observation that such questions can be argued against.
    \item \textbf{A reusable data-collection method.} The domain-labeled occurrence sets produced at the occurrence-localization stage (Section~\ref{sec:pipeline-locate}), namely a bridge form's occurrences grouped by source-document domain at Wikipedia scale and without manual annotation, are themselves a resource independent of any downstream analysis, and can seed further sense-annotation, evaluation, or teaching material.
\end{itemize}
Realizing any of these uses at scale, across more models, more bridge forms, or with statistical validation, is beyond what a single toolkit run or this manual undertakes; Section~\ref{sec:limitations} states that boundary explicitly.

\section{Background and Related Work}
\label{sec:background}

\subsection{Classical Foundations}
\label{sec:background-classical}
Static distributional embeddings such as word2vec and GloVe assign exactly one vector per word type, trained so that words appearing in similar contexts end up nearby: a direct implementation of the distributional hypothesis that a word is known by the company it keeps \citep{firth1957synopsis, mikolov2013efficient, pennington2014glove}. Under such a representation, a polysemous or homonymous word's senses are necessarily conflated into a single point, weighted by their relative frequency in the training corpus; no mechanism exists to separate them per occurrence.\parbreak
Contextual representations, beginning with ELMo \citep{peters2018elmo} and continuing through transformer encoders and decoders built on the attention-based architecture of \citet{vaswani2017attention}, replaced the one-vector-per-type assumption with a function from a token \emph{and its surrounding context} to a vector \citep{devlin2019bert}. The expectation that such models individuate word senses by context is widely held and is consistent with their downstream performance, but is usually supported indirectly: through probing classifiers that recover linguistic properties from hidden states \citep{tenney2019bert}, or through performance gains on word-sense disambiguation and related tasks. \citet{belinkov2019analysis} survey this broader analysis literature and note that a probing classifier's success shows only that some property is linearly decodable from a representation, not that the network itself relies on that property downstream. This toolkit turns on a related but distinct question, one the survey does not itself address: not whether a property is decodable by a trained probe, but whether a model's own geometry spontaneously separates senses with no decoder in the loop at all.\parbreak
Word-sense disambiguation benchmarks typically rely on manually sense-tagged corpora, which are accurate but limited in scale and coverage, and which necessarily commit to a particular sense inventory (e.g. WordNet-style sense splits). The bridge-form design used by this toolkit takes a different, complementary route: rather than annotating individual occurrences with a sense label, it treats \emph{topical domain of the source document} as a naturalistic, low-cost proxy for sense, obtained at scale from encyclopedic category structure rather than from manual annotation. This trades sense-level precision (a domain-typical document may occasionally use a bridge form in an atypical sense) for scale and for the specific structural guarantee described in Section~\ref{sec:intro}: because the proxy is external to the model and to the word's identity, any resulting separation in a model's representation space is attributable to context rather than to an artifact of the labeling procedure itself.\parbreak
Measuring the resulting separation is itself a design question. Cluster-validity indices such as the silhouette coefficient \citep{rousseeuw1987silhouettes} give a bounded, interpretable score for how well a labeling matches the geometry of a representation space, and are common in representation analysis; Section~\ref{sec:design-pairwise} discusses why this toolkit restricts silhouette computation to one domain pair at a time rather than pooling three or more domains, which is not the coefficient's default use.

\subsection{Recent Work (2024--2026)}
\label{sec:background-recent}
The premise that transformer language models individuate word senses by context has been examined from several angles in the last three years, none of which take the specific route this toolkit takes. Four clusters of recent work are relevant.

\subsubsection{Behavioral word-sense disambiguation with LLMs}
\label{sec:behavioral_word-sense}
A first line of work asks whether an LLM's \emph{output}, a sense label, a definition, a generated example, is correct, not how its internal representations are organized. \citet{meconi2025wordsenses} test instruction-tuned LLMs on word-sense disambiguation framed both as classification and as generation (producing a definition or example for a target sense), and find that leading LLMs match specialized WSD systems and generalize better across domains, with generative framings performing especially well.\parbreak
\citet{basile2025wsdllm} extend a large multilingual WSD benchmark with LLM-oriented definition-generation and definition-selection subtasks and find strong zero-shot LLM performance, though fine-tuned mid-sized models still win on the original classification framing.\parbreak
\citet{navigli2026wsddead}, in an invited overview, argues that word-sense disambiguation is not obsolete in the LLM era but is better understood as a diagnostic lens on an LLM's lexical-semantic competence, pointing to persistent failures on infrequent senses. This cluster is informative about what a model \emph{says}; it is silent on whether the geometry behind that output already separates senses without a classifier or a generation head making the decision, which is the question this toolkit's silhouette measurement (Section~\ref{sec:pipeline-measure}) asks directly.

\subsubsection{Unsupervised clustering of contextual embeddings against gold sense inventories}
\label{sec:unsupervised_clustering}
A second line clusters contextual embeddings to induce or evaluate word senses, but validates the clustering against manually sense-tagged resources. \citet{mosolova2025wsi} revisit word-sense induction evaluation, arguing that prior benchmarks use unrealistic sense distributions, and find that no fully unsupervised clustering method (nor an LLM asked to cluster) beats a naive one-cluster-per-lemma baseline on their SemCor-derived data.\parbreak
\citet{lietard2024concepts} induce per-lemma senses from BERT embeddings with agglomerative and $k$-means clustering, then cluster senses further into cross-lexicon concepts, evaluating against SemCor with (extended) BCubed F1. Both use a gold sense inventory as ground truth and standard cluster-quality metrics (not the silhouette coefficient, and not a pairwise protocol); neither addresses the pooled-silhouette bias this toolkit's pairwise design is built to avoid (Section~\ref{sec:design-pairwise}), because neither uses a silhouette-style index at all. The bridge-form design instead treats naturally occurring \emph{topical domain} as an unlabeled, low-cost sense proxy, trading gold-standard precision for scale and for the absence of a sense inventory to commit to in advance.

\subsubsection{Layer-wise probing and representation geometry}
\label{sec:layer-wise_probing}
A third line asks where in a network's depth particular information is encoded. \citet{liu2024fantastic} probe Llama-2's 32 layers with the Word-in-Context task and anisotropy-corrected cosine similarity, finding lexical semantics peaks around layers 6--8 in that decoder; in encoder models, by contrast, later layers are typically more informative.\parbreak
\citet{skean2025layerbylayer} propose a cross-architecture (autoregressive, bidirectional, and state-space) information-theoretic framework for layer-wise representation quality, evaluated on general embedding benchmarks rather than on lexical ambiguity specifically, and find that training objective, not modality, governs whether representation quality is concentrated mid-network or distributed more uniformly across layers.\parbreak
A recent systematic review \citep{lopezotal2025interpretability} surveys some 160 papers on what transformer layers encode, including a subsection on lexical semantics and polysemy, and can be read as a map of this broader literature. None of these studies is built as a reusable toolkit spanning both bidirectional and causal architectures for a single, targeted question: the role this manuscript's pipeline is designed to fill (Section~\ref{sec:design-architecture}).

\subsubsection{Ambiguity-focused geometric and causal analyses}
\label{sec:ambiguity-focused_analyses}
Closest in spirit to the bridge-form construct are studies that analyze layer-wise geometry specifically for ambiguous words. \citet{riviere2025sawc} build a Spanish minimal-pair dataset of ambiguous nouns with human relatedness judgments and correlate layer-wise cosine distances from BERT-family encoders against those judgments, asking which layers best track human-perceived sense distinctions; their ground truth is graded human judgment, not a domain corpus, and only encoder models are tested.\parbreak
Most directly relevant, and contemporaneous with this manuscript, \citet{scott2026divergent} analyze homonym and polyseme representations layer by layer in three causal decoders (GPT-2-Small, Llama-3.2-3B, Qwen2.5-32B), and report that representations of an ambiguous word's two senses become maximally distinct in \emph{middle} layers and then \emph{reconverge} by the final layers even as the models' output predictions keep diverging. They establish this pattern causally using activation patching and layer ablation, on curated ambiguous word pairs rather than a multi-domain corpus. That finding is a direct, and useful, caution for any method, this toolkit included, that reads late-layer cosine geometry as the natural place to look for sense separation. Section~\ref{sec:pipeline-visualize} and Section~\ref{sec:interpreting} therefore report the toolkit's silhouette curves and PCA panels across \emph{every} layer, not only the final one, precisely so that a middle-layer peak followed by later convergence, were it to occur for a given model and bridge form, would be visible in the output rather than hidden by only inspecting hs[-1].

\subsubsection{Where do the bridge forms stand?}
\label{sec:bridge_forms_stand}
Across these four clusters, no study combines (i) an unlabeled, naturalistic, domain-based proxy for word sense in place of a manually sense-tagged inventory, (ii) a domain-pairwise silhouette protocol that avoids the pooled-silhouette bias of standard multi-group clustering validity indices, and (iii) a single reusable pipeline spanning both bidirectional encoders and causal decoders. The closest work is either behavioral (asking what a model outputs, not how its representations are organized), evaluated against gold sense annotations, focused on general-purpose layer-wise information content rather than lexical ambiguity specifically, or, in the case of the most closely related contemporaneous study \citep{scott2026divergent}, decoder-only and causal-interventionist rather than an unsupervised geometric-clustering design applied across architecture families. This is the specific gap the toolkit documented in this manual is built to fill, and Section~\ref{sec:limitations} states plainly what the toolkit still does not do.

\section{Design Principles}
\label{sec:design}
This section states the toolkit's central design choices together with the specific methodological failure mode each one is meant to prevent. Implementation detail is deferred to Section~\ref{sec:toolkit}.

\subsection{Fix the form, vary the context}
\label{sec:design-form}
A bridge form's defining property is that its written form is held fixed while its context (and, correspondingly, its intended sense) varies across domains. This is what licenses the interpretation of any layer-wise separation as a context effect: at the embedding layer, every occurrence of the same word type maps to the same vector up to positional encoding, regardless of which domain it comes from. If separation is absent at that layer and present at later ones, the later-layer separation cannot be attributed to the word type itself; it has to come from what surrounds it. This is a property of the construct, verifiable directly from the model's embedding layer, not an assumption that has to be taken on trust.

\subsection{Specific, non-overlapping category labels}
\label{sec:design-categories}
Each domain a bridge form bridges is mapped to a specific source category, not a broad one. A broad label (e.g. a generic ``Geography'' category) mixes senses that the construct is supposed to keep apart, economic geography alongside physical geography for a form like \emph{current}, and contaminates a domain's occurrence pool with instances that do not carry the sense the domain is meant to isolate. Category choice is fixed \emph{before} any representation is extracted or any score is computed: a bridge form is only informative if what counts as one occurrence of a given sense is decided in advance, not selected after seeing which choice produces a cleaner separation.

\subsection{Pairwise, not pooled, comparison}
\label{sec:design-pairwise}
When three or more domains are pooled into a single silhouette computation, the between-group term $b(i)$ for a given point is defined as the mean distance to the \emph{nearest} other group only. This means that if two domains happen to sit close together in a model's representation geometry, they drag each other's score down, while a third domain that happens to be geometrically isolated looks artificially clean by comparison: an artifact of how many other groups are in the pool, not a fact about that domain's separation from any one of them. The toolkit avoids this by computing silhouette scores one domain pair at a time (Section~\ref{sec:pipeline-measure}): every reported score concerns exactly two domains, and is not affected by the presence or absence of a third.

\subsection{Full representation width, not a low-dimensional projection}
\label{sec:design-width}
Separation is measured at the model's full hidden width, in cosine distance, never on a two-dimensional projection. A two-dimensional view is useful for illustration but is, by construction, a lossy summary of a much higher-dimensional space; a projection that looks well separated (or poorly separated) is not guaranteed to reflect the geometry of the full space it was derived from. The toolkit keeps these roles distinct: the silhouette score is what is reported as evidence, computed at full width; the paired PCA scatter (Section~\ref{sec:pipeline-visualize}) illustrates, but does not substitute for, that score.

\subsection{A shared basis for comparable layers}
\label{sec:design-shared-pca}
When two layers (the embedding layer and the model's final layer) are each visualized as a two-dimensional scatter, the toolkit fits a single PCA basis \citep{jolliffe2002pca} on both layers' points pooled together, then projects each layer onto that shared basis, rather than fitting an independent PCA per layer. Two independent projections would each show their own best-case, independently optimized view of their respective layer; any apparent ``movement'' of the point cloud between the two panels could then be, at least in part, an artifact of the axes being re-oriented rather than a real displacement of the points. A shared basis treats both layers symmetrically and keeps the two panels' axes directly comparable, at the cost of not showing either layer at its own best two-dimensional resolution: a deliberate trade against a specific interpretive risk.

\subsection{Offset-based, not string-based, token localization}
\label{sec:design-offsets}
A bridge form's own token is located in a tokenized sequence using the tokenizer's character-offset mapping, matching the character span of the word occurrence in the original text back to a token index, rather than by matching subword strings. Byte-level BPE vocabularies, used by several of the model families this toolkit supports, can spell the same word differently depending on whether it is word-initial or occurs mid-sequence; string matching against subword tokens is therefore unreliable in a way that offset matching, tied to the original character positions, is not.

\subsection{Architecture-neutral design, stated caveats}
\label{sec:design-architecture}
The pipeline treats a model as a black box that returns one hidden-state vector per layer per token, which makes it applicable, without modification, to both bidirectional encoders (e.g. BERT-family models) and causal decoders (e.g. GPT-2-family models). The two families differ in a way that matters for this specific construct: a token's representation in a bidirectional encoder can draw on the entire input sequence, while a token's representation in a causal decoder can only draw on what precedes it. A bridge form occurring early in a sentence therefore has structurally less context available to it under a causal model. This is a property of the architecture being tested, not a defect of the toolkit, but it means that a lower separation score under a causal model does not, on its own, support a claim of weaker context-sensitivity without controlling for the form's position in the sentence: a caveat the toolkit surfaces (Section~\ref{sec:implementation}) rather than resolves automatically.

\section{The Toolkit}
\label{sec:toolkit}
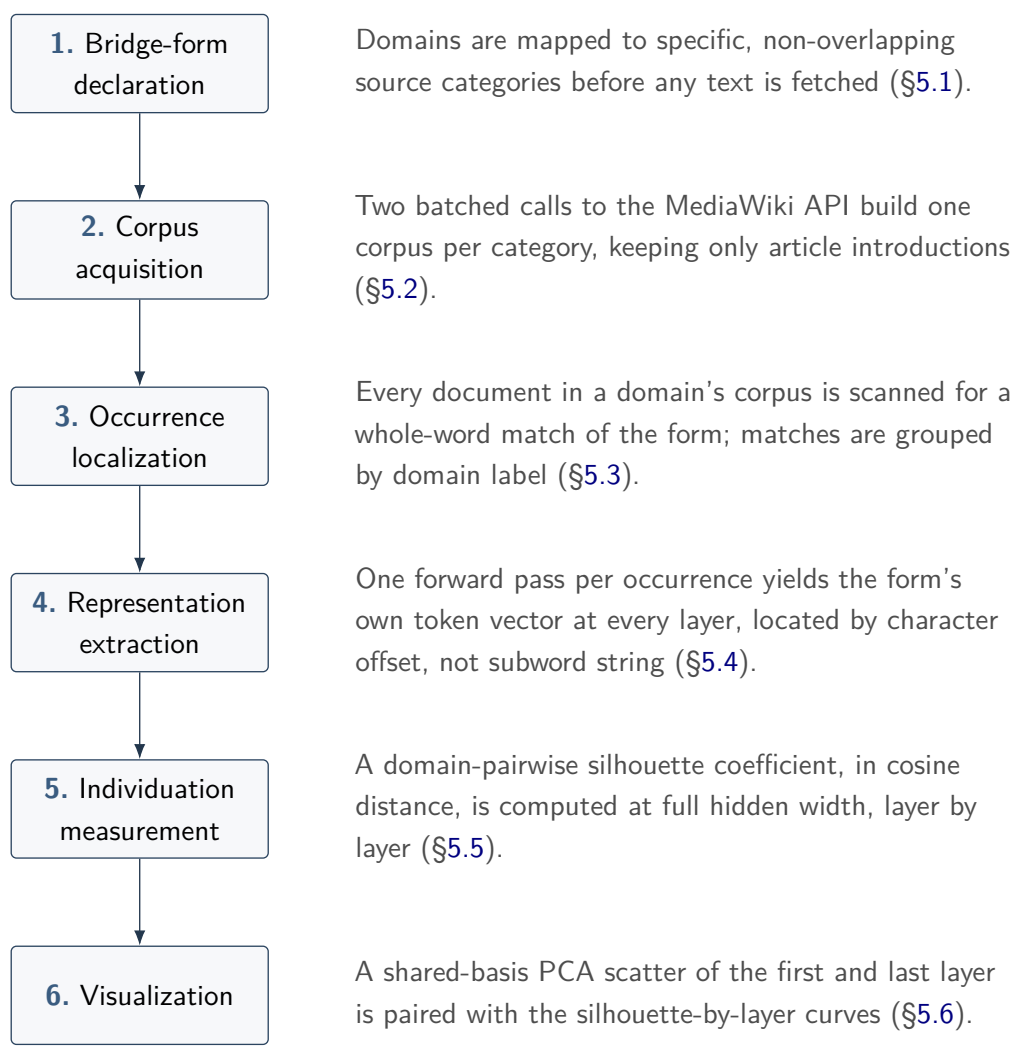
\begin{figure}[H]
\centering
\definecolor{pipe}{RGB}{58,92,128} 

\begin{tikzpicture}[
  node distance=1.15cm,
  stage/.style={
    rectangle, rounded corners=2pt, draw=pipe!75!black, line width=0.5pt,
    fill=pipe!4,
    minimum width=3.4cm, minimum height=1.3cm,
    align=center, font=\small, text width=3.1cm
  },
  comment/.style={
    align=left, font=\small, text=black!70, text width=8.7cm
  },
  arr/.style={-{Latex[length=2mm]}, line width=0.5pt, pipe!60!black}
]
  \node[stage] (s1) {\textcolor{pipe}{\bfseries 1.}\ Bridge-form declaration};
  \node[stage, below=of s1] (s2) {\textcolor{pipe}{\bfseries 2.}\ Corpus acquisition};
  \node[stage, below=of s2] (s3) {\textcolor{pipe}{\bfseries 3.}\ Occurrence localization};
  \node[stage, below=of s3] (s4) {\textcolor{pipe}{\bfseries 4.}\ Representation extraction};
  \node[stage, below=of s4] (s5) {\textcolor{pipe}{\bfseries 5.}\ Individuation measurement};
  \node[stage, below=of s5] (s6) {\textcolor{pipe}{\bfseries 6.}\ Visualization};

  \draw[arr] (s1) -- (s2);
  \draw[arr] (s2) -- (s3);
  \draw[arr] (s3) -- (s4);
  \draw[arr] (s4) -- (s5);
  \draw[arr] (s5) -- (s6);

  \node[comment, right=1.0cm of s1] (c1)
    {Domains are mapped to specific, non-overlapping source categories before any text is fetched (\S\ref{sec:pipeline-declare}).};
  \node[comment, right=1.0cm of s2] (c2)
    {Two batched calls to the MediaWiki API build one corpus per category, keeping only article introductions (\S\ref{sec:pipeline-corpus}).};
  \node[comment, right=1.0cm of s3] (c3)
    {Every document in a domain's corpus is scanned for a whole-word match of the form; matches are grouped by domain label (\S\ref{sec:pipeline-locate}).};
  \node[comment, right=1.0cm of s4] (c4)
    {One forward pass per occurrence yields the form's own token vector at every layer, located by character offset, not subword string (\S\ref{sec:pipeline-extract}).};
  \node[comment, right=1.0cm of s5] (c5)
    {A domain-pairwise silhouette coefficient, in cosine distance, is computed at full hidden width, layer by layer (\S\ref{sec:pipeline-measure}).};
  \node[comment, right=1.0cm of s6] (c6)
    {A shared-basis PCA scatter of the first and last layer is paired with the silhouette-by-layer curves (\S\ref{sec:pipeline-visualize}).};
\end{tikzpicture}
\caption{The six pipeline stages, top to bottom; each stage's output is the next stage's input. The comment beside each box summarizes what happens at that stage and why; full detail is in the correspondingly numbered subsection of this section.}
\label{fig:pipeline}
\end{figure}

The pipeline has six stages, summarized in Figure~\ref{fig:pipeline} and described individually below. Each stage consumes the output of the one before it and can, in principle, be re-run or substituted independently (a different corpus source in place of Wikipedia, for instance) without requiring changes to the later stages, provided the same interface, a mapping from document identifiers to plain text, is respected.

\setstretch{1.15}

\setstretch{1.5}
\FloatBarrier
\subsection{Bridge-form declaration}
\label{sec:pipeline-declare}
Every bridge form is declared as a mapping from a domain label (e.g. \texttt{"physics"}) to the specific source category that supplies that domain's text (e.g. \texttt{"Electricity"}). This declaration is the toolkit's single source of truth for what counts as an occurrence of a given sense (Section~\ref{sec:design-categories}): it is authored once, before any text is fetched or any vector is extracted, and every later stage is driven by it rather than by a choice made after inspecting intermediate results. Listing~\ref{lst:declare} reproduces the opening entries of this declaration exactly as authored in the toolkit's reference notebook (reformatted here from its original single-line-per-form layout for legibility only; no key, value, or entry is altered).\parbreak

\setstretch{1.15}
\begin{lstlisting}[caption={Opening entries of the bridge-form declaration, from the toolkit's reference notebook.}, label={lst:declare}]
BRIDGE_FORMS = {
    "current": {
        "physics":   "Electricity",
        "economics": "Finance",
        "geography": "Physical geography",
    },
    "bank": {
        "economics": "Banking",
        "geography": "Rivers",
    },
    # further bridge forms (field, mass, wave, charge, cell, ...)
    # follow the same schema; see the reference notebook for the
    # full, currently active list
}
\end{lstlisting}
\setstretch{1.5}

\subsection{Corpus acquisition}
\label{sec:pipeline-corpus}
For every distinct source category referenced anywhere in the declaration, fetched once regardless of how many bridge forms or domains share it, the toolkit builds a small corpus of article introductions via two calls to the MediaWiki Action API \citep{mediawiki_api}:
\begin{enumerate}
  \item \textbf{Category membership.} The article titles directly in the category, plus the titles of its direct subcategories, one level deep. A Wikipedia category by itself typically holds few articles directly; most usable volume sits one level down, in subcategories, and going exactly one level down is a large gain in coverage without walking the full category tree (and without risking topic drift from descending further).
  \item \textbf{Introduction extracts.} The plain-text introduction only of each article: Wikipedia's own version of a comparable, abstract-length unit, chosen so that corpus composition is not skewed toward whichever article happens to be longest.
\end{enumerate}

A documented, non-obvious API behavior governs how the second call is batched: requesting a full-text extract for a batch of titles caps the number of pages that actually receive an extract at one per request, regardless of batch size; requesting the introduction only is what allows a batch of up to twenty titles to return twenty extracts. Batching without this setting silently truncates a corpus to a small, non-representative fraction of the intended titles: a failure mode that does not raise an error and is only visible by checking yield per category. As API etiquette, requests identify the tool and a contact address via the \texttt{User-Agent} header, read from a local, untracked configuration file rather than hard-coded. \parbreak
Listing~\ref{lst:extracts} reproduces the toolkit's own implementation of the second call, unmodified from its reference notebook; the \texttt{exintro=True} parameter is exactly the non-obvious setting just described, and \texttt{explaintext=True} is what returns plain text rather than the article's wiki markup. \parbreak

\setstretch{1.15}
\begin{lstlisting}[caption={Batched introduction-extract fetch, from the toolkit's reference notebook.}, label={lst:extracts}]
def fetch_extracts(titles, headers, batch_size=20):
    """Plain-text introduction of each article title, in batches.

    `exintro=True` is required for a batch to return more than one extract -
    without it, the API silently returns a full-article extract for only the
    first title of each batch and empty extracts for the rest.
    """
    url = 'https://en.wikipedia.org/w/api.php'
    extracts = {}

    for i in range(0, len(titles), batch_size):
        batch = titles[i:i + batch_size]
        params = {
            'action': 'query',
            'prop': 'extracts',
            'titles': '|'.join(batch),
            'explaintext': True,
            'exintro': True,
            'exlimit': 20,
            'format': 'json',
        }
        response = requests.get(
            url, params=params, headers=headers, timeout=20)
        response.raise_for_status()
        pages = response.json()['query']['pages']

        for page in pages.values():
            title = page.get('title')
            extract = page.get('extract', '')
            if title:
                extracts[title] = extract

    return extracts
\end{lstlisting}
\setstretch{1.5}

\subsection{Occurrence localization}
\label{sec:pipeline-locate}
Within each domain's own corpus, every document is searched for a whole-word, case-insensitive match of the bridge form (a word-boundary-delimited regular expression). Matches are grouped by \emph{domain label}, not by source category name: the label is what carries meaning downstream, as the group a silhouette computation compares; the category is only the means by which the corresponding text was sourced, and is not itself compared against anything. Listing~\ref{lst:search} reproduces the toolkit's own implementation, unmodified from its reference notebook. \parbreak

\setstretch{1.15}
\begin{lstlisting}[caption={Occurrence localization, grouped by domain label, from the toolkit's reference notebook.}, label={lst:search}]
def search_bridge_forms(corpus, bridge_forms):
    """Article titles containing each bridge form, grouped by subject area.

    Returns {form: {topic_label: [titles]}}.
    """
    results = {}

    for form, topics in bridge_forms.items():
        pattern = re.compile(rf'\b{re.escape(form)}\b', re.IGNORECASE)
        results[form] = {}

        for topic, category in topics.items():
            matches = [
                title for title, text in corpus.get(category, {}).items()
                if pattern.search(text)
            ]
            results[form][topic] = matches

    return results
\end{lstlisting}
\setstretch{1.5}

\subsection{Representation extraction}
\label{sec:pipeline-extract}
For each located occurrence, the document is tokenized once, with character-offset mapping enabled, and passed through the model once with hidden-state output enabled, so that a single forward pass yields the token's vector at \emph{every} layer simultaneously: the embedding-layer output plus one output per transformer block. The bridge form's own token position is found via the offset mapping (Section~\ref{sec:design-offsets}): the token whose character span contains the start of the regex match. \parbreak
Two constraints keep the extracted sample well defined rather than silently degraded. Only the \emph{first} occurrence of the form per document is kept, so that no single long document can contribute a disproportionate share of one domain's points. And if the located character match falls outside the tokenizer's truncation window, the occurrence is dropped rather than misaligned to the wrong token: an explicit exclusion rather than a silent corruption of the sample. \parbreak
Listing~\ref{lst:extract} reproduces the toolkit's own implementation, unmodified from its reference notebook. \texttt{pattern.search(text)} returns only the first match, which is what enforces the first-occurrence-only constraint; \texttt{token\_idx is None} is exactly the truncation-window exclusion just described, implemented via the offset-based localization of Section~\ref{sec:design-offsets}; and the loop over \texttt{outputs.hidden\_states} is the single forward pass yielding every layer's vector at once. \parbreak

\setstretch{1.15}
\begin{lstlisting}[caption={Layer-wise representation extraction, from the toolkit's reference notebook.}, label={lst:extract}]
def extract_bridge_vectors(bridge_matches, bridge_forms, corpus, tokenizer, model, max_length=512):
    """One vector per layer, per occurrence, for every form in `bridge_matches`.

    Returns {form: {"layers": [layer_0_X, ..., layer_N_X], "group": labels, "doc_id": titles}}.
    """
    model.eval()
    results = {}

    for form, topic_matches in bridge_matches.items():
        pattern = re.compile(rf'\b{re.escape(form)}\b', re.IGNORECASE)
        layer_vectors, groups, doc_ids = None, [], []

        for topic, titles in topic_matches.items():
            category = bridge_forms[form][topic]
            for title in titles:
                text = corpus[category][title]

                match = pattern.search(text)
                if not match:
                    continue

                encoding = tokenizer(
                    text, return_tensors='pt', return_offsets_mapping=True,
                    truncation=True, max_length=max_length,
                )
                offsets = encoding.pop('offset_mapping')[0].tolist()

                token_idx = next(
                    (i for i, (start, end) in enumerate(offsets)
                     if start <= match.start() < end),
                    None,
                )
                if token_idx is None:
                    continue  # match fell outside the truncated window

                with torch.no_grad():
                    outputs = model(**encoding)

                if layer_vectors is None:
                    layer_vectors = [[] for _ in outputs.hidden_states]
                for layer_idx, layer_hs in enumerate(outputs.hidden_states):
                    layer_vectors[layer_idx].append(layer_hs[0, token_idx].numpy())

                groups.append(topic)
                doc_ids.append(title)

        if layer_vectors is None:
            continue

        results[form] = {
            "layers": [np.stack(v) for v in layer_vectors],
            "group": np.array(groups),
            "doc_id": np.array(doc_ids),
        }

    return results
\end{lstlisting}

\setstretch{1.5}
\subsection{Individuation measurement}
\label{sec:pipeline-measure}
For a bridge form with domains $\{d_1, d_2, \ldots\}$, the toolkit computes, independently for every unordered pair $(d_i, d_j)$ and for every layer $\ell$, the silhouette coefficient in cosine distance restricted to the occurrences belonging to that pair:

\begin{equation}
s(i) \;=\; \frac{b(i) - a(i)}{\max\bigl(a(i), b(i)\bigr)},
\label{eq:silhouette}
\end{equation}

where $a(i)$ is the mean cosine distance from point $i$ to the other points in its own domain, and $b(i)$ is the mean cosine distance from point $i$ to the points of the other domain in the pair (Section~\ref{sec:design-pairwise} explains why this is computed pair by pair rather than with all domains pooled). The reported quantity per pair, per layer, is the mean of $s(i)$ over all points in that pair. At the embedding layer, this score is expected, by the structural argument of Section~\ref{sec:design-form}, to sit near zero for every pair; how it behaves at later layers is an empirical question this manual does not answer (Section~\ref{sec:interpreting}). \parbreak
Distance is measured in cosine, not Euclidean, terms. A hidden-state vector's norm reflects factors that are not, by themselves, evidence about lexical sense: token frequency, sequence position, and layer-normalization scaling all affect a vector's magnitude without necessarily tracking meaning. Cosine distance discards magnitude and retains only the angle between two vectors, which is the standard corrective for this in the layer-wise probing literature: \citet{liu2024fantastic}, for instance, adopt an anisotropy-corrected cosine similarity for the same reason when probing lexical semantics layer by layer. The silhouette coefficient is itself metric-agnostic \citep{rousseeuw1987silhouettes}, so this is a choice the toolkit makes deliberately rather than one the coefficient forces on it. \parbreak
Listing~\ref{lst:silhouette} reproduces the toolkit's own implementation of this computation, unmodified from its reference notebook, called once per bridge form and reused across the pairwise loop of Section~\ref{sec:design-pairwise} (\texttt{silhouette\_score} is \texttt{sklearn.metrics.silhouette\_score} and \texttt{combinations} is \texttt{itertools.combinations}). Here \texttt{layers} is a list holding one $N \times d$ matrix per model layer, and \texttt{group} is the aligned array of domain labels described in Section~\ref{sec:pipeline-locate}; restricting the comparison to \texttt{mask}, one domain pair at a time, is exactly the pairwise-not-pooled design justified in Section~\ref{sec:design-pairwise}. \parbreak

\setstretch{1.15}
\begin{lstlisting}[caption={Pairwise, layer-wise silhouette computation, from the toolkit's reference notebook.}, label={lst:silhouette}]
def silhouette_pairwise_by_layer(bridge_vectors, form):
    """Silhouette per layer, for each PAIR of subject areas, computed separately.

    With 3+ areas pooled together, b(i) is the distance to the nearest OTHER
    group only - two areas that happen to sit close to each other drag each
    other's score down while an isolated third area looks artificially clean.
    Restricting to one pair at a time removes that ambiguity.
    """
    layers = bridge_vectors[form]["layers"]
    group = bridge_vectors[form]["group"]
    topics = np.unique(group)

    results = {}
    for topic_a, topic_b in combinations(topics, 2):
        mask = np.isin(group, [topic_a, topic_b])
        scores = [
            silhouette_score(X[mask], group[mask], metric='cosine')
            for X in layers
        ]
        results[(topic_a, topic_b)] = scores

    return results
\end{lstlisting}
\setstretch{1.5}

\subsection{Visualization}
\label{sec:pipeline-visualize}
For each bridge form, the toolkit produces a single figure with three panels: a paired PCA scatter of the embedding layer and the final layer (Section~\ref{sec:design-shared-pca}), on shared, symmetric axis limits and a shared aspect ratio so that a visible contraction or spread of the point cloud between the two panels reflects the underlying geometry rather than independent auto-scaling; and the pairwise silhouette-by-layer curves of Section~\ref{sec:pipeline-measure}, one curve per domain pair. Each domain is assigned both a distinct marker shape and a distinct color, so that domain identity remains legible under grayscale reproduction or projection, and each pairwise silhouette curve is colored as a blend of its two domains' colors, so that which pair a curve belongs to is identifiable from color alone. Listing~\ref{lst:project} reproduces the toolkit's own implementation of the shared-basis projection just described (Section~\ref{sec:design-shared-pca}), unmodified from its reference notebook; the full plotting routine calls this once per bridge form, then draws the two PCA panels and the pairwise silhouette-by-layer panel from its output and from Listing~\ref{lst:silhouette}, respectively. \parbreak

\setstretch{1.15}
\begin{lstlisting}[caption={Shared-basis PCA projection, from the toolkit's reference notebook.}, label={lst:project}]
def project_shared(X_list):
    """Fit one PCA on all of `X_list` pooled, then project each item onto it.

    A shared basis keeps every projected layer on the same two axes, so
    panels can be compared directly instead of each showing its own
    independently-optimised view.
    """
    p = PCA(n_components=2, random_state=0)
    p.fit(np.concatenate(X_list, axis=0))
    return [p.transform(X) for X in X_list], p.explained_variance_ratio_
\end{lstlisting}
\setstretch{1.5}

\begin{figure}[H]
\centering
\setstretch{1.15}
\begin{subfigure}[b]{\textwidth}
  \centering
  \includegraphics[width=\textwidth]{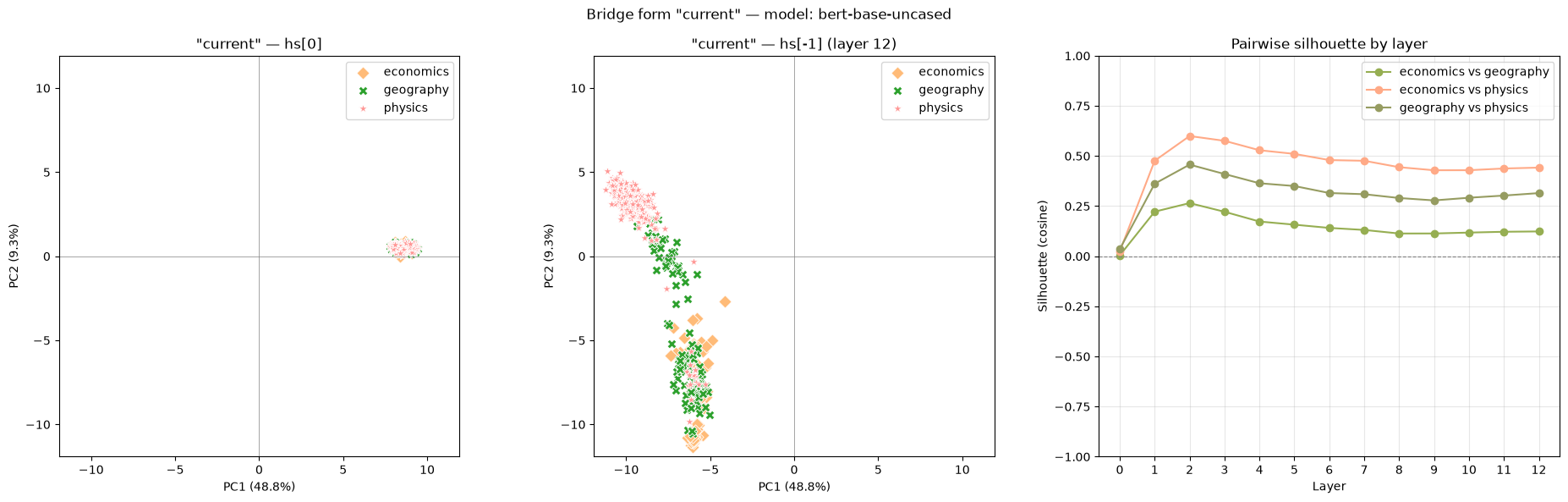}
  \caption{Bridge form \emph{current}, three domains.}
  \label{fig:example-current}
\end{subfigure}
\vspace{8pt}
\begin{subfigure}[b]{\textwidth}
  \centering
  \includegraphics[width=\textwidth]{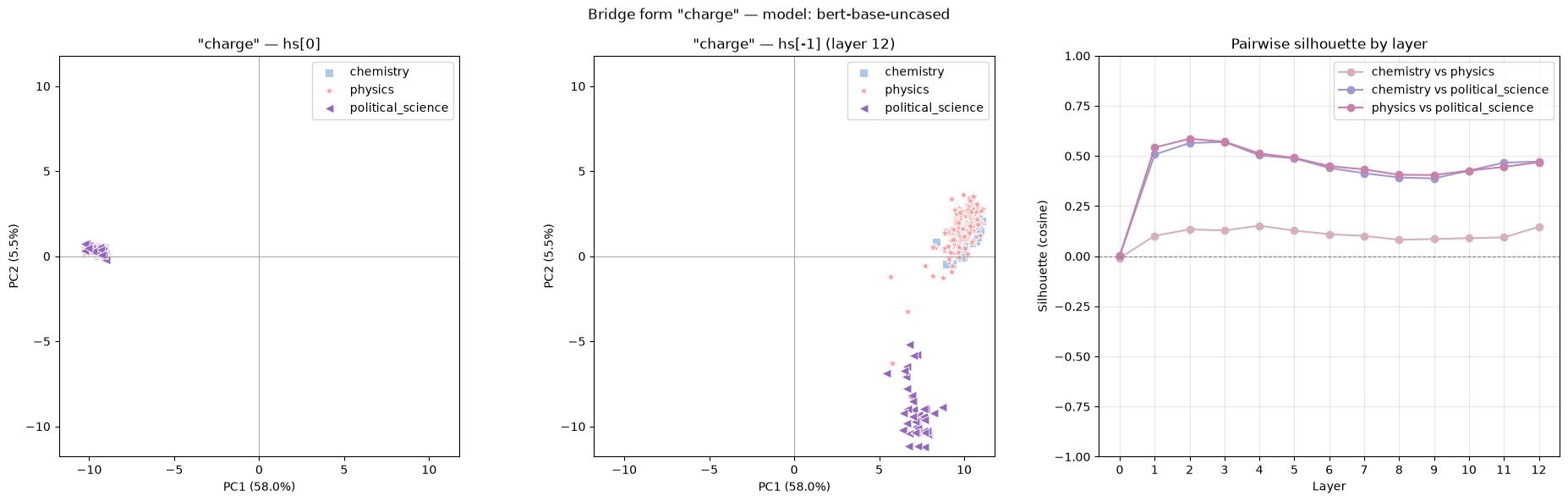}
  \caption{Bridge form \emph{charge}, three domains.}
  \label{fig:example-charge}
\end{subfigure}
\caption{Two examples of the toolkit's standard output: the paired PCA scatter (embedding layer, left; final layer, center) and the pairwise silhouette-by-layer curves (right). These two runs are shown \emph{only} to illustrate the output format described in this section: panel layout, marker/color encoding, and axis conventions. The second example also happens to illustrate that the three pairwise curves in a panel need not coincide: two of them run close together across the layer range while the third stays visibly apart, which is the kind of pattern this display is built to make legible. Consistent with the scope stated in Section~\ref{sec:intro}, no explanation for why any particular pair behaves as it does is offered here; the specific pattern each panel happens to show is not reported, interpreted, or generalized as a finding of this manuscript.}
\label{fig:example-output}
\end{figure}

Figure~\ref{fig:example-output} shows two examples of the output this pipeline stage produces; Section~\ref{sec:interpreting} explains how each panel is meant to be read.

\setstretch{1.5}
\FloatBarrier

\section{Implementation and Reproducibility}
\label{sec:implementation}

\subsection{Supported model families}
\label{sec:supported_model_families}
The toolkit has been exercised against the architectures in Table~\ref{tab:models}, chosen to span bidirectional encoders and causal decoders, and encoder variants that differ in pretraining objective, size, and attention mechanism. Any model exposing a standard hidden-states-per-layer interface (as implemented by the Hugging Face \texttt{transformers} library, \citealp{wolf2020transformers}) can be substituted without modification to the pipeline.

\setstretch{1.15}
\begin{table}[H]
\centering
\small
\begin{tabular}{@{}llll@{}}
\toprule
Model & Architecture family & Context direction & Reference \\
\midrule
\texttt{bert-base-uncased} & Encoder & Bidirectional & \citep{devlin2019bert} \\
\texttt{roberta-base} & Encoder & Bidirectional & \citep{liu2019roberta} \\
\texttt{distilbert-base-uncased} & Encoder (distilled) & Bidirectional & \citep{sanh2019distilbert} \\
\texttt{microsoft/deberta-base} & Encoder & Bidirectional & \citep{he2021deberta} \\
\texttt{gpt2} & Decoder & Causal & \citep{radford2019gpt2} \\
\texttt{EleutherAI/pythia-410m} & Decoder & Causal & \citep{biderman2023pythia} \\
\bottomrule
\end{tabular}
\caption{Model families the toolkit's interface has been designed and checked against. The pipeline processes one model per run; comparing architectures means re-running the full pipeline with a different selection.}
\label{tab:models}
\end{table}

\setstretch{1.5}
\FloatBarrier

\subsection{Software dependencies}
\label{sec:software_dependencies}
The reference implementation is a single, sequentially executed notebook using \texttt{transformers} and \texttt{torch} for model loading and inference, \texttt{scikit-learn} for the silhouette coefficient and PCA \citep{pedregosa2011scikitlearn}, and \texttt{matplotlib}/\texttt{numpy} for visualization and array handling. All dependencies are pinned to specific versions in the accompanying environment file archived with the toolkit (Section~\ref{sec:availability}).

\subsection{Determinism}
\label{sec:determinism}
PCA fitting uses a fixed random seed; corpus acquisition and model inference are otherwise deterministic given a fixed corpus snapshot and model checkpoint, with the caveat that Wikipedia category contents can change between runs (Section~\ref{sec:limitations}) and that the corpus a given run operates on should therefore be archived alongside any results derived from it, not re-fetched on demand. \parbreak
\textbf{One model per run.} By design, a single execution of the pipeline processes exactly one model, selected from a maintained list, given the computational cost of running several transformer models end to end. Comparing architectures is a matter of re-running the full pipeline with a different selection, not a built-in batch operation: a deliberate scope boundary, discussed further in Section~\ref{sec:limitations}.

\subsection{Data provenance and etiquette}
\label{sec:data_provenience}
All corpus text is fetched live from the public Wikipedia API at run time, identified by a descriptive \texttt{User-Agent} and contact address (Section~\ref{sec:pipeline-corpus}) read from a local, untracked configuration file, in line with the API's request etiquette; no Wikipedia text is redistributed as part of the toolkit's own archive beyond what its corresponding license permits.

\section{Interpreting the Toolkit's Output}
\label{sec:interpreting}
This section describes what each element of the toolkit's output is designed to show and how it is intended to be read, illustrated with the two examples of the output format shown as Figure~\ref{fig:example-output} in Section~\ref{sec:pipeline-visualize}. It is written independently of any specific claim about what a model does: it is a reading guide for the instrument, not a report of what the instrument found for these particular examples. \parbreak
The \textbf{embedding-layer panel} (left panel of Figure~\ref{fig:example-output}) is expected, by the structural argument of Section~\ref{sec:design-form}, to show no domain-related structure: every occurrence of a bridge form should project to close to the same region, regardless of domain, because the embedding layer's output for a fixed word type varies only with position, not with context. This panel functions as a manipulation check on the construct itself, prior to any claim about later layers. If it fails to show this collapse, the bridge form's occurrences are not, in fact, sharing the same vocabulary entry as assumed, and the construct's structural guarantee does not hold for that case. \parbreak
The \textbf{final-layer panel} (center) and the \textbf{pairwise silhouette-by-layer curves} (right) are where any domain-related organization, if present, becomes visible. A silhouette curve that departs from zero indicates that the two domains in that pair are, at that layer and in that representation space, separable to a degree quantified by the coefficient of Equation~\eqref{eq:silhouette}; a curve's shape across layers, whether separation appears early and is sustained, appears late, or is not sustained, is itself part of what such a curve is meant to convey, independent of its value at any single layer. The percentage reported alongside each PCA axis is the share of \emph{full-hidden-width} variance that axis captures, typically a small number in a space of several hundred dimensions; it quantifies how much of the geometry the two-dimensional panel is showing, not how separated the domains are: that claim is carried by the silhouette score computed at full width (Section~\ref{sec:design-width}), not by the scatter plot. \parbreak
This manual does not present or interpret the specific outcome of running the toolkit on any given model or bridge-form set: what pattern the panels and curves take for a particular model, bridge form, or layer range is an empirical question, and is the subject of work that uses this toolkit, not of this manuscript.

\section{Scope and Limitations}
\label{sec:limitations}
The following are explicit boundaries of the current toolkit, stated so that they can be addressed by future work rather than mistaken for claims the toolkit does make.
\begin{itemize}[nosep]
  \item \textbf{No built-in significance testing.} Silhouette scores are reported as point estimates per layer, per domain pair. The toolkit does not currently implement a permutation baseline or a confidence interval around these scores; a reader (or a subsequent study using the toolkit) should treat a reported score's departure from zero in light of the number of occurrences behind it, which the toolkit does report but does not itself gate on.
  \item \textbf{No minimum-sample-size enforcement.} Domain occurrence counts depend on how frequently a bridge form appears in that domain's corpus and can be small; the toolkit computes and plots a silhouette curve regardless of how few occurrences support it, leaving sample-size adequacy to the user.
  \item \textbf{Coverage depends on Wikipedia category structure.} A declared source category may hold few or no directly-attributed articles beyond what its one-level-deep subcategories supply (Section~\ref{sec:pipeline-corpus}), and category structure changes over time; corpus size and stability are therefore bounded by an external, evolving resource rather than by the toolkit itself.
  \item \textbf{Domain label as a proxy for sense, not a verified annotation.} Grouping occurrences by source-document domain (Section~\ref{sec:design-categories}) is a scalable proxy for sense, not a per-occurrence sense annotation; a domain-typical document can, in principle, use a bridge form in an atypical sense, which the toolkit does not detect or filter.
  \item \textbf{One model per run, by design.} Cross-architecture comparison (Section~\ref{sec:implementation}) requires re-running the pipeline once per model and is not automated within a single run.
  \item \textbf{First-occurrence-only sampling.} Only a document's first occurrence of a bridge form is kept (Section~\ref{sec:pipeline-extract}), which bounds sample size below the total number of occurrences available in the corpus.
\end{itemize}

None of the above compromises the structural guarantee described in Section~\ref{sec:design-form}, which concerns the embedding layer and does not depend on sample size, category coverage, or which model is selected. They bound what can be concluded from a \emph{particular} run's later-layer results, which is precisely the kind of claim this manuscript, by design, does not make (Section~\ref{sec:interpreting}).

\section{Data and Code Availability}
\label{sec:availability}
The toolkit's full source code, the environment specification needed to reproduce its software dependencies, and the corpus snapshot(s) used to exercise it are archived under a persistent identifier, independently of this manuscript:
\begin{quote}
\bibentry{toolkit_zenodo}
\end{quote}
The toolkit is released under a Creative Commons Attribution 4.0 International (CC BY 4.0) license. This manuscript itself is intended for archival on arXiv under a a Creative Commons
Attribution 4.0 International (CC BY 4.0) license. Please cite the archived toolkit (not this manuscript alone) when reporting results produced with it; please cite this manuscript when referring to its design rationale.

\section{Conclusion}
\label{sec:conclusion}
The bridge-form construct gives a structurally clean way to ask whether a transformer language model individuates a word's occurrences by context: because the construct fixes word form while varying domain-appropriate sense, any separation that appears beyond the embedding layer cannot be attributed to word identity. The toolkit documented here operationalizes that construct end to end, and this manual has stated, and justified, the design choices that make its output interpretable: pairwise rather than pooled comparison, full-width rather than projected measurement, offset-based token localization, and a shared basis for visually comparable layers, among others. What the instrument reads out when applied to specific models and bridge forms is a separate, empirical question, addressed in work that builds on it rather than in this manuscript.

\section*{Author Contributions}
All authors contributed to the conception and design of the toolkit described in this manuscript, and to writing and reviewing the manuscript.

\section*{Acknowledgments}
J.L.V.M. gratefully acknowledges the postdoctoral fellowship received under the Postdoctoral Program in Research Management (PPDG), Office of the Pro-Rector for Research (PRP), University of Campinas (Unicamp), administered through the Center for Energy and Petroleum Studies (CEPETRO).\parbreak
The authors thank the Power Electronics Laboratories (LEPO) and the Center for Electric Mobility Research (CEMOBE) for institutional and infrastructural support throughout this work.\parbreak
The authors thank Lucas Hideki Ueda for the arXiv endorsement of this submission.\parbreak
The authors declare no conflict of interest.

\setstretch{1.15}
\bibliographystyle{apalike}
\bibliography{references}
\end{document}